\documentclass[conference]{IEEEtran}
\IEEEoverridecommandlockouts
\usepackage{cite}
\usepackage{amsmath,amssymb,amsfonts}
\usepackage{amstext}
\usepackage{mathtools}
\usepackage{algorithmic}
\usepackage{graphicx}
\usepackage{textcomp}
\usepackage{xcolor}
\usepackage{float}
\usepackage{stackengine}
\usepackage{booktabs}
\usepackage{tabularx}
\usepackage{caption}
\usepackage{subcaption}
\usepackage{textcomp}
\usepackage{placeins}
\usepackage{hyperref}
\usepackage[first=0,last=9]{lcg}

\usepackage{hhline}
\usepackage{colortbl}
\definecolor{Gray}{gray}{0.85}
\def\BibTeX{{\rm B\kern-.05em{\sc i\kern-.025em b}\kern-.08em
    T\kern-.1667em\lower.7ex\hbox{E}\kern-.125emX}}
    
\begin{document}

\newcommand{\R}{\mathbb{R}}

\title{Evaluation of Deep Neural Network Domain Adaptation Techniques for Image Recognition\\
\thanks{*Equal contribution.}
}

\author{\IEEEauthorblockN{Venkata Santosh Sai Ramireddy Muthireddy\textbf{*}}
\IEEEauthorblockA{\textit{Hochschule Bonn-Rhein-Sieg}\\
Bonn, Germany \\
santosh.muthireddy@smail.inf.h-brs.de}
\and
\IEEEauthorblockN{Alan Preciado-Grijalva\textbf{*}}
\IEEEauthorblockA{\textit{Hochschule Bonn-Rhein-Sieg}\\
Bonn, Germany \\
alan.preciado@smail.inf.h-brs.de}
}

\maketitle

\begin{abstract}
It has been well proved that deep networks are efficient at extracting features from a given (source) labeled dataset. However, it is not always the case that they can generalize well to other (target) datasets which very often have a different underlying distribution. In this report, we evaluate four different domain adaptation techniques for image classification tasks: DeepCORAL, DeepDomainConfusion, CDAN and CDAN+E. These techniques are unsupervised given that the target dataset dopes not carry any labels during training phase. We evaluate model performance on the office-31 dataset. A link to the github repository of this report can be found here: \href{https://github.com/agrija9/Deep-Unsupervised-Domain-Adaptation}{https://github.com/agrija9/Deep-Unsupervised-Domain-Adaptation}. 
\end{abstract}


\section*{Introduction}
\begin{figure}[H]
    \centering
    \includegraphics[scale=0.4]{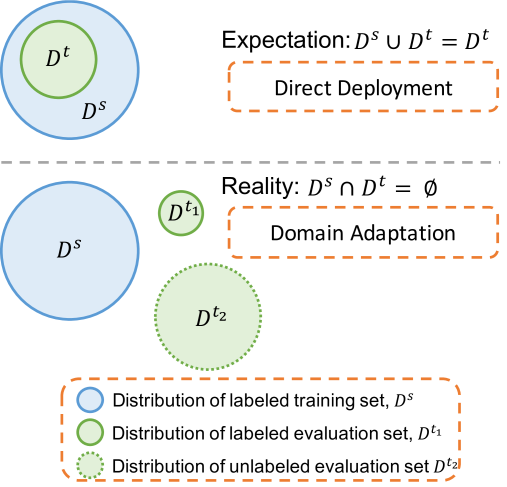}
    \caption{Domain adaptation in the true data space: Expectation vs. Reality.\cite{8953559}}
    \label{fig:expVSreal}
\end{figure}

Deep neural networks based methods have been producing state-of-the-art (SOTA) results for many problems in machine learning and computer vision. These methods require large amount of training and testing data to achieve the expected result. Although the model is trained with large datasets sometimes it will not generalize learned knowledge to new environment and datasets. This is because deep learning algorithms assume that training and testing data is drawn from independent and identical distributions (i.i.d.). However this assumption rarely holds true, as there will be a shift in data distributions across different domains this is explained in Fig. \ref{fig:expVSreal}. This domain shift between source and target datasets will make deep neural networks produce wrong predictions on the target dataset. So training of a deep neural network with source and target datasets which reduces the domain shift between distribution of datasets is called as \textbf{domain adaptation} \cite{5288526}.

There are different types of domain adaptation (DA) techniques, a few of them are \textit{Unsupervised DA} \cite{NIPS2018_7436}\cite{10.1007/978-3-319-49409-8_35}, \textit{Semi Supervised DA}\cite{9010425}, \textit{Weakly Supervised DA}\cite{8954126}, \textit{One Shot DA}\cite{hoffman2013one}, \textit{Few Shot DA}\cite{8953674}, \textit{Zero Shot DA}\cite{10.1007/978-3-030-01252-6_47}. In this report, we discussed on Unsupervised DA techniques. Unsupervised DA is called so because unlabeled target dataset is used during training to reduce the domain shift. 

Further Unsupervised DA have different types based on the technique used to reduce the domain shift between the source and target datasets. They are:
\begin{itemize}
    \item Adversarial Methods \cite{NIPS2018_7436}
    \item Distance-based Methods \cite{10.1007/978-3-319-49409-8_35} \cite{tzeng2014deep}
    \item Incremental Methods \cite{8460982}
    \item Optimal Transport \cite{jimenez2019cdot}
    \item Other Methods \cite{Binkowski_2019_ICCV}
\end{itemize}

In this report, techniques evaluated are selected from Adversarial and Distance-based methods. In Adversarial methods, Conditional adversarial domain  adaptation (CDAN) and CDAN with Entropy conditioning (CDAN+E) \cite{NIPS2018_7436} are selected. In Distance-based methods, Deep domain  confusion:  Maximizing  for  domain  invariance (DDC) \cite{tzeng2014deep} and Deep  coral:  Correlation  alignment  for  deep domain  adaptation \cite{10.1007/978-3-319-49409-8_35} are selected. Application of domain adaptation techniques can be used in different types of domain shift explained in Fig. \ref{fig:diff_type_domain_shift} and application in dataset to dataset domain shift is selected for evaluation. In this report evaluation of different domain adaptation techniques are benchmarked on office-31 dataset \cite{10.1007/978-3-642-15561-1_16}.

\begin{figure}[H]
    \centering
    \includegraphics[scale=0.15]{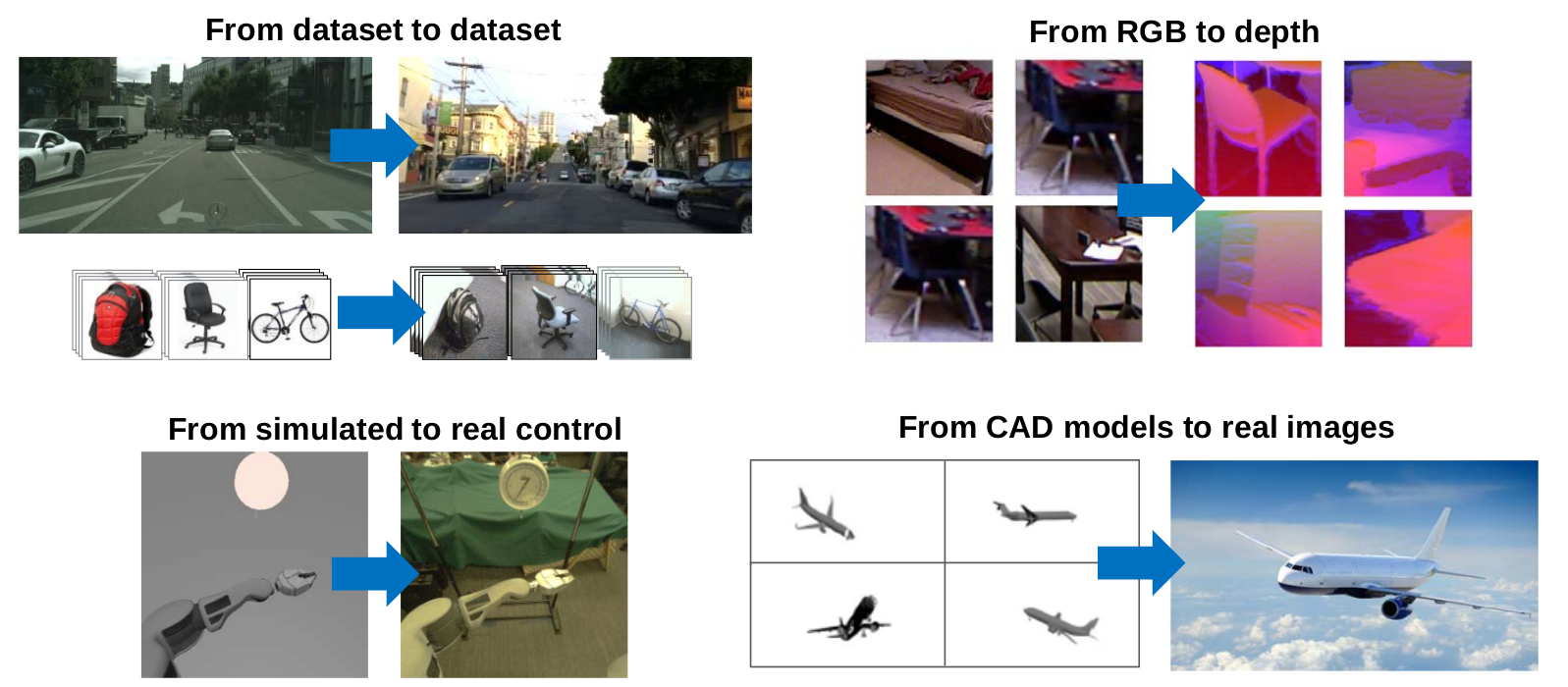}
    \caption{Applications to different types of domain shift \cite{5995702}}
    \label{fig:diff_type_domain_shift}
\end{figure}

%%%%%%%%%%%%%%%%%%%%%%%%%%%%%%%%%%%%%%%%
\section*{Domain Adaptation Methods}

\subsection*{DeepCORAL}

DeepCORAL \cite{10.1007/978-3-319-49409-8_35} is a domain adaptation method based on a previously proposed method called Correlation Alignment (CORAL) \cite{DBLP:journals/corr/SunFS16}. 

CORAL attempts to increase performance in domain shift by aligning the second order statistics between both data distributions. This method consists of 1) feature extraction, 2) apply a linear transformation and 3) implement an SVM classifier. 

DeepCORAL extends the capacities of CORAL by learning a non-linear mapping using convolutional neural networks and a differentiable loss function that attempts to minimize the Frobenius norm between the feature covariance matrices of source and target distributions. This method works in an unsupervised way where the target dataset has no labels. 

Recall that the main goal is to learn representations that allow to achieve good performance in both data distributions (image recognition, for example). To achieve this, DeepCORAL intializes the network parameters from pre-trained AlexNet model while at the same time minimizing the loss mentioned above (\textit{a.k.a} CORAL loss). The architecture of DeepCORAL is shown in the figure below. 

\begin{figure}[H]
    \centering
    \includegraphics[scale=0.30]{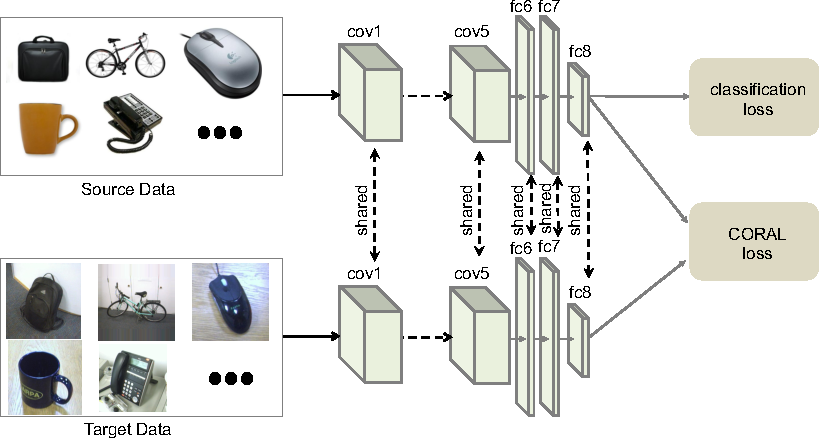}
    \caption{DeepCORAL neural network architecture.}
    \label{fig:diff_type_domain_shift}
\end{figure}

Note that DeepCORAL takes the last fully connected layer of AlexNet (\textit{fc8}) and computes a classification and CORAL loss. 

\subsubsection*{CORAL loss}
This loss is defined for a single layer feature map. In this case for \textit{fc8}, although not mentioned, we believe the authors run a grid search to identify the optimal layer within the network to compute the loss. This can have to do with the task at hand. 

To formalize the CORAL loss, suppose that we have source-domain training examples $D_s = \{x_i\}$, where $x \in \R^d$ with labels $L_s = \{y_i\}$, $i \in \{1,...,L\}$, and unlabeled data $D_T = \{u_i\}$, $u \in \R^d$. Let $n_s$ and $n_t$ be the number of source and target data available respectively. Recall that \textbf{x} and \textbf{u} are the d-dimensional deep layer activations (\textit{fc8}) $\phi{(I)}$ of an input image $I$. Let $C_S$ and $C_T$ be the feature covariance matrices of these activations for source and target domains respectively. The CORAL loss is defined as the difference between these covariances (second order statistics):

\begin{equation}
    \textit{l}_{CORAL} = \frac{1}{4d^2} \|C_S - C_T\|_F^2
\end{equation}

where $\|\cdot\|_F^2$ refers to the squared matrix Frobenius norm. The covariance matrices $C_S$, $C_T$ are defined as

\begin{equation}
   C_S = \frac{1}{n_S - 1}\big(D_S^T D_S - \frac{1}{n_S}(1^T D_S)^T(1^T D_S)\big)
\end{equation}

\begin{equation}
   C_T = \frac{1}{n_T - 1}\big(D_T^T D_T - \frac{1}{n_T}(1^T D_T)^T(1^T D_T)\big)
\end{equation}

where \textbf{1} is a column vector with all elements equal to 1.

The gradient of the CORAL loss can be calculated using the chain rule. Refer to the paper for the analytical expression. 

In practice, we do batch learning and thus the covariances are computed over batches. Moreover, the network parameters are shared between the two networks. 

In the context of multi-class classification, the total loss that is used to train is a joint term between with classification loss (cross-entropy) and CORAL loss. The first one trying to optimize classification accuracy and the second trying to learn features that can work well in both data domains. The total loss is 

\begin{equation}
   \textit{l} = l_{CLASS} + \sum_{i=1}^{t}{\lambda_i l_{CORAL}}
\end{equation}

here $t$ is the number of CORAL loss computed per layers. In this case it is a single layer. The parameter $\lambda$ is introduced as a regularizer to help reach an equilibrium between these two losses at the end of training. 

%%%%%%%%%%%%%%%%%%%%%%%%%%%%%%%%%%%%%%
\subsection*{Deep Domain Confusion (DDC)}

Deep domain confusion is another domain adaptation method that uses a deep neural network to learn a non-linear transformation. It is similar to DeepCORAL in the sense that it introduces a transfer loss to learn features for both data domains. \cite{tzeng2014deep}

In their paper, Tzeng et al. show the concept of biased datasets and how trained classifiers don't always transfer well to target domains. Their approach is to maximize a loss function that minimizes classification error and maximizes domain confusion.

\begin{figure}[H]
    \centering
    \includegraphics[scale=0.65]{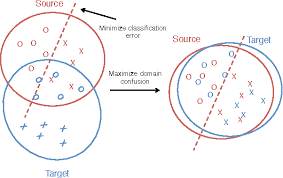}
    \caption{Deep Domain Confusion approach: Maximizing domain confusion can lead to good performance in target domains.}
    \label{fig:diff_type_domain_shift}
\end{figure}

To minimize the distance between data distributions, DDC uses a standard distribution distance metric called Maximum Mean Discrepancy (MMD). In this case, this distance is computed with respect to an activation layer $\phi(\cdot)$ acting on source data points $x_S$ $\in$ $X_S$ and target data points $x_T$ $\in$ $X_T$. The approximation to this distance is as follows

\begin{equation}
    MMD(X_S, X_T) = \left| \frac{1}{\|X_S\|}\sum{\phi(x_s)} - \frac{1}{\|X_T\|}\sum{\phi(x_t)} \right|
\end{equation}

Training a model with MMD and a classification loss will result in a strong classifier ready to transfer accross domains. Thus, the total loss is

\begin{equation}
  l = l_C(X_L,y) + \lambda MMD^2(X_S,X_T)
\end{equation}

Here $l_C$ corresponds to a classification loss on the available source labelled data $X_L$ and the ground truths $y$. The hyperparameter $\lambda$ acts in a similar way a regularizer as in DeepCORAL.

The architecture proposed in DDC is similar to DeepCORAL; it intializes its weights with a pre-trained AlexNet and weights are shared between both data domains. The difference is that it has an extra \textit{bottleneck} layer \textit{fc adapt} that is meant to regularize the training of the source classifier and prevent overfitting to particular nuances of the source distribution. The loss is computed after this bottleneck layer. The next figure shows the architecture of DDC.

\begin{figure}[H]
    \centering
    \includegraphics[scale=0.90]{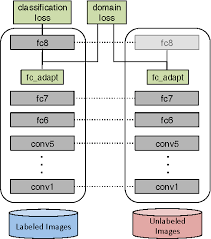}
    \caption{Deep Domain Confusion neural network architecture. Both data domains share weights, a bottleneck is introduced to the AlexNet architecture and the total loss is computed after this layer.}
    \label{fig:diff_type_domain_shift}
\end{figure}

\begin{figure}[H]
    \centering
    \includegraphics[scale=0.90]{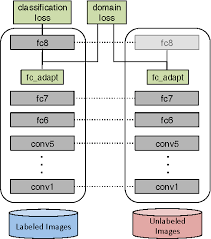}
    \caption{Deep Domain Confusion approach: Maximizing domain confusion can lead to good performance in target domains.}
    \label{fig:diff_type_domain_shift}
\end{figure}

%%%%%%%%%%%%%%%%%%%%%%%%%%%%%%%%%%%%%%
\subsection*{CDAN}
Adversarial domain adaptation \cite{pmlr-v37-ganin15} \cite{Tzeng_2017_CVPR} \cite{Tzeng_2015_ICCV} consists of domain adaptation and adversarial learning which is very similar to Generative Adversarial Networks (GANs) \cite{goodfellow2014generative}. In adversarial domain adaptation, a deep neural network model learns image representations by minimizing a classification loss and simultaneously a domain discriminator learns how to distinguish between source and target domain. There are however two pitfalls for adversarial domain adaptation methods. In the first place they cannot handle datasets which have complex multi modal distributions. And secondly we cannot rely on the uncertain information difference between source and target data to model the domain discriminator.  

These two pitfalls can be solved by conditional domain adaptation techniques. Conditional domain adaptation is made possible due to the research carried out in Conditional Generative Adversarial Networks (CGANs) \cite{mirza2014conditional}. CGANs proved a point that domain shift between distributions of real and synthetic images can be reduced by conditioning network using discriminative information. 

CDAN \cite{NIPS2018_7436} is an adversarial domain adaptation technique motivated by CGANs and it makes use of the discriminative information obtained from a deep classifier network. The key to CDAN models is a new conditional domain discriminator conditioned to the covariance of representations of domain-specific resources and classifier predictions. Let us consider $D_s = \{(x_i^s,y_i^s)\}_{i=1}^{n_s}$ is a source domain data with $n_s$ labeled examples and $D_t = \{(x_j^t)\}_{j=1}^{n_t}$ is a target domain dataset with $n_t$ unlabeled examples. Source and target datasets are drawn from $P(x^s,y^s)$ and $Q(x^t,y^t)$ which are not i.i.d. i.e., $P\ne Q$. The aim of CDAN is to design a network $G: x \to y$ that can reduce the domain shift between source and target datasets. $f=F(x)$ and $g=G(x)$ are feature representation and classifier predictions made from the network G respectively.

Considering all the things mentioned above, CDAN is formalized as a minmax optimization problem. In minmax optimization, two terms are considered as players, namely $E(G)$ corresponding to source classifier G and minimized to guarantee lower source risk and $E(D,G)$ on G and the domain discriminator D on cross domains. $E(D,G)$ is minimized over D and maximized over $f$ and $g$

\begin{equation}
    E(G) = \mathbb{E}_{(x_i^s,y_i^s)\sim D_s}L(G(x_i^s),y_i^s)
\end{equation}
\begin{equation}
    E(D,G) = -\mathbb{E}_{x_i^s\sim D_s}log[D(f_i^s,g_i^s)]-\mathbb{E}_{x_i^t\sim D_t}log[D(f_j^t,g_j^t)]
\end{equation}
where L(.) is the cross entropy loss. Now CDAN is described in minmax optimization problem as 
\begin{equation}
    \stackunder{min}{G} \ E(G) - \lambda E(D,G) \ and \
    \stackunder{min}{G} \ E(D,G)
\end{equation}
where $\lambda$ is a hyper-parameter that balances the domain adversary and source risk.
\begin{figure}[H]
    \centering
    \includegraphics[scale=0.4]{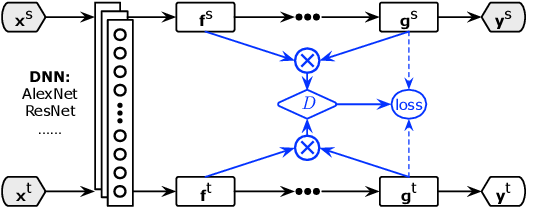}
    \caption{Architecture of CDAN for domain adaptation \cite{NIPS2018_7436}}
    \label{fig:cdan_architecture}
\end{figure}
In Fig. \ref{fig:cdan_architecture} D is the domain discriminator conditioned using a multilinear map $f \otimes g$, multilinear mapping is explained in Section 3.2 of \cite{NIPS2018_7436}. Thus CDAN enables domain adaptation over cross domains using minmax optimization. 
\subsection*{CDAN+E}\label{sec:cdan+e}
CDAN+E is an variant of CDAN. In CDAN+E we will have additional involvement of \textbf{Entropy Conditioning} which improves the difference between source and target domains. In CDAN, the domain discriminator allots equal weight to each and every example irrespective of examples with uncertain predictions from G. So uncertainity corresponding to classifier predictions are quantified using entropy criteria explained in Eq. \ref{eq:entropy}. 
\begin{equation}
\label{eq:entropy}
    H(g) = -\sum_{c=1}^C g_c log \ g_c
\end{equation}
where C is number of classes and $g_c$ is probability of predictions.

Now training examples are prioritized by an entropy-aware weight given by
\begin{equation}
    w(H(g)) = 1+e^{-H(g)}
\end{equation}
So CDAN+E minmax optimization is formulated as 
\begin{equation}
\begin{multlined}[0.5\linewidth]
    \stackunder{min}{G}\mathbb{E}_{(x_i^s,y_i^s)\sim D_s}L(G(x_i^s),y_i^s) \\ +
    \lambda (\mathbb{E}_{x_i \sim D_s}w(H(g_i^s))log[D(T(h_i^s))] \\ +\mathbb{E}_{x_j^t \sim D_t}w(H(g_j^t))log[1-D(T(h_j^t))]) 
\\
\stackunder{max}{D}\mathbb{E}_{x_i \sim D_s}w(H(g_i^s))log[D(T(h_i^s))] \\ +\mathbb{E}_{x_j^t \sim D_t}w(H(g_j^t))log[1-D(T(h_j^t))]
\end{multlined}
\end{equation}
where h =(f,g) is a joint variable.

Thus CDAN+E encourages the certain predictions which are achieved through entropy minimization principle \cite{grandvalet2005semi}.
%%%%%%%%%%%%%%%%%%%%%%%%%%%%%%%%%%%%%%%%
\section*{Experiments}

In this section we describe the experiments performed with the different metods on different data domains. Our implementations are written in Pytorch and are located in the link to the repository given above. The experiments were carried out using Google Colab GPU support (Testla P4).

\subsection*{Office-31 dataset}

  Office-31 dataset was first introduced in \cite{10.1007/978-3-642-15561-1_16}. It contains a total of 4652 images collected from three different sources. The three different sources from which images are collected are online web (amazon), digital SLR (dslr) camera and webcam. Office-31 dataset have 31 classes \footnote{The 31 classes are: backpack, bike, bike helmet, bookcase, bottle, calculator, desk chair, desk lamp, computer, file cabinet, headphones, keyboard, laptop, letter tray, mobile phone, monitor, mouse, mug, notebook, pen, phone, printer, projector, puncher, ring binder, ruler, scissors, speaker, stapler, tape, and trash can.} in total.
  
  \begin{itemize}
      \item \textbf{Images from the web:} These are collected from amazon website. We call this dataset  as amazon dataset (A). It has around 2800 images and with an average of 90 images per each class. The resolution of the images are each  300x300 pixels. 
      \item \textbf{Images from a dslr camera:} These are captured from dslr camera in real environments.  We call this dataset  as dslr dataset (D). It has total of 498 images and with an average of 16 images per each class. The resolution of the images are 1000x1000 pixels. 
      \item \textbf{Images from a webcam:} These are collected from a webcam with lower resolution and will have more noise. From now this is called as webcam dataset (W). It has around 795 images and with an average of 25 images per each class. The  resolution of the images is HxW where H and W are varying between 400 and 600. 
  \end{itemize}
  \begin{figure}[H]
      \centering
      \includegraphics[scale=0.7]{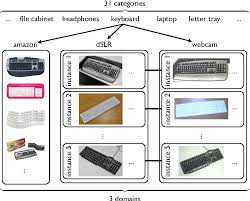}
      \caption{Sample of keyboard category images for all the three domains \cite{10.1007/978-3-642-15561-1_16}}
      \label{fig:office_dataset_keybord}
  \end{figure}
%   \item Pytorch framework

%%%%%%%%%%%%%%%%%%%%%%%%%%%%%%%%%%%%%%%%%
\subsection*{DeepCORAL}

The method is evaluated with the benchmark Office dataset described above. We have carried out 6 different experiments for all the domain shifts available (given the tree datastes cotained in Office). 

CORAL loss is applied to the last fully connected layer (fc8). The weights of this layer are initialized following a normal distribution $\mathcal{N}(0, 0.005)$ and the weights from the other layers are initialized with the pre-trained AlexNet. We used Stochastic gradient descent (SGD).

DeepCORAL was trained using batches of 128 images (both domains), a fixed learning rate of $10^{-3}$, weight decay set to $5x10^{-4}$ and momentum to 0.9. The $\lambda$ factor is set in such a way that both losses are approximately the same. In practice, given $t$ epoch iterations, $\lambda$ is $\frac{1}{t}$. Furthermore, during training only the source dataset is labelled, this is how we compute the classification loss. 

In Table 1 we show the obtained accuracies for each method for all 6 experiments. These values correspond to the image classification accuracy on the target dataset. From here, we see that DeepCORAL obtained the best result in 4 out of 6 experiments. This method outputs accuracy values in the range between 38\% to 89\%. The fluctuation of accuracies between domains has to do directly with the chosen source and target distributions. It is important to note that the method does not output accuracies greater than 50\% in most of the cases. We've compared our values with the original paper and although lower by a small percentage, these follow the same trend as the original results.

For a better visualization of the method's performance we have added to visualize accuracy and loss as a function of epochs for the case Amazon $\,\to\,$ Webcam. Figure 8a) shows an increase of almost 10\% in recognition accuracy when including CORAL loss. Figure 9a) shows the behavior of the loss functions, note that they tend to become equal due to the lambda factor.

%%%%%%%%%%%%%%%%%%%%%%%%%%%%%%%%%%%%%%%%%
\subsection*{Deep Domain Confusion}

Training for this method is very similar to DeepCORAL. The main differences are 1) it trains with MMD loss, 2) adds the bottleneck layer and 3) it uses a schedule learning rate which decreases as a function of epochs. Fig. 8b) shows that accuracy with MMD (or DDC) doesn't have a very noticeable increase compared to training without MMD, however, it seems to be more stable and shows an steady increase. Due to time constrains, we were only able to run one experiment for this method. However, it is consistent with the accuracy reported on the original paper.

%%%%%%%%%%%%%%%%%%%%%%%%%%%%%%%%%%%%%%%%%
\subsection*{CDAN and CDAN+E}

Experiments for CDAN and CDAN+E are conducted with same set of hyperparameters and the only difference between both is the transfer loss function that is involved. Hyperparameters used in the experiments for CDAN and CDAN+E are:
\begin{itemize}
    \item Epochs 100
    \item Source batch size 10
    \item Target batch size 10
    \item Learning rate 1e-3
    \item Momentum 0.9
    \item Weight decay 5e-4
\end{itemize}

Dataset used for all the experiments are office-31 dataset \cite{10.1007/978-3-642-15561-1_16} Backbone used in this experiment is pretrained AlexNet with a small modification. There is a bottleneck layer introduced before last fully connected layer. Bottleneck layer has the dimension 256 (used in the original paper \cite{NIPS2018_7436}) which is also a user defined shape. Optimizer used is Stochastic Gradient Descent (SGD) in which parameters of deep neural network G and domain discriminator D are added to optimize. During the training phase each batch from source and target data loader are passed through deep neural network G and features f from bottleneck layer and fully connected layer output are obtained from the deep neural network G. Thus obtained features f and final layer output of source and target datasets are joined together. Concatenated final layer outputs are passed through softmax layer in order to get network predictions g. 

\begin{figure}[]
\begin{subfigure}{.5\linewidth}
\centering
\includegraphics[scale=0.15]{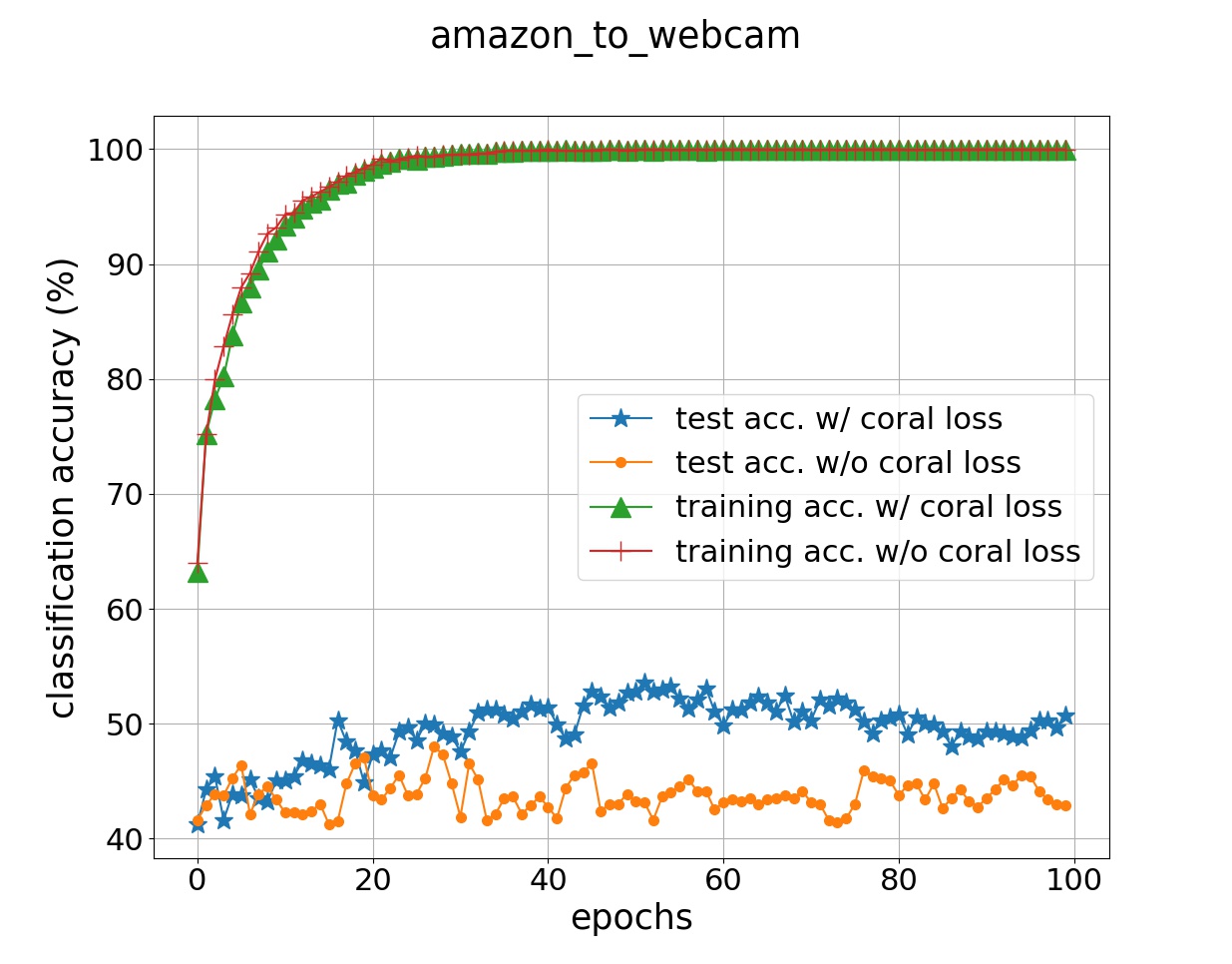}
\caption{DeepCORAL}
\label{fig:deepcoral_acc}
\end{subfigure}%
\begin{subfigure}{.5\linewidth}
\centering
\includegraphics[scale=0.25]{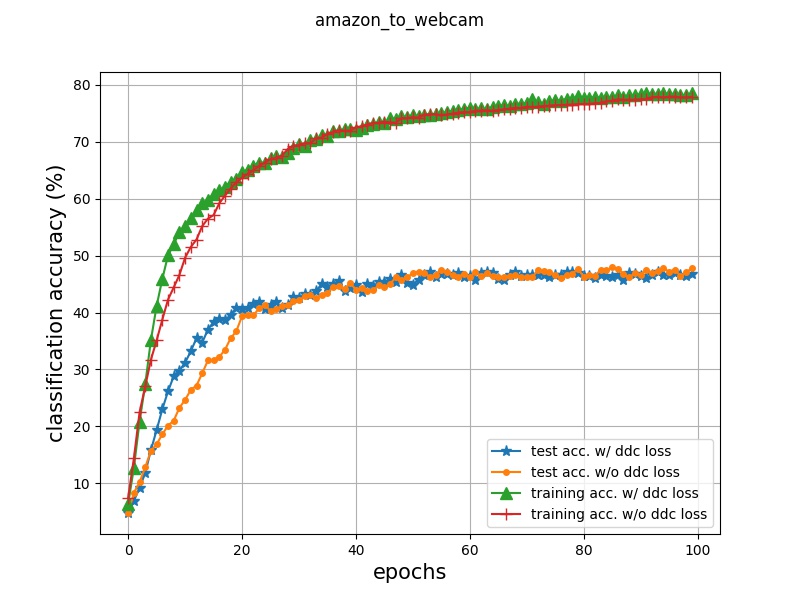}
\caption{DDC}
\label{fig:ddc_acc}
\end{subfigure}\\[1ex]
\begin{subfigure}{.5\linewidth}
\centering
\includegraphics[scale=0.25]{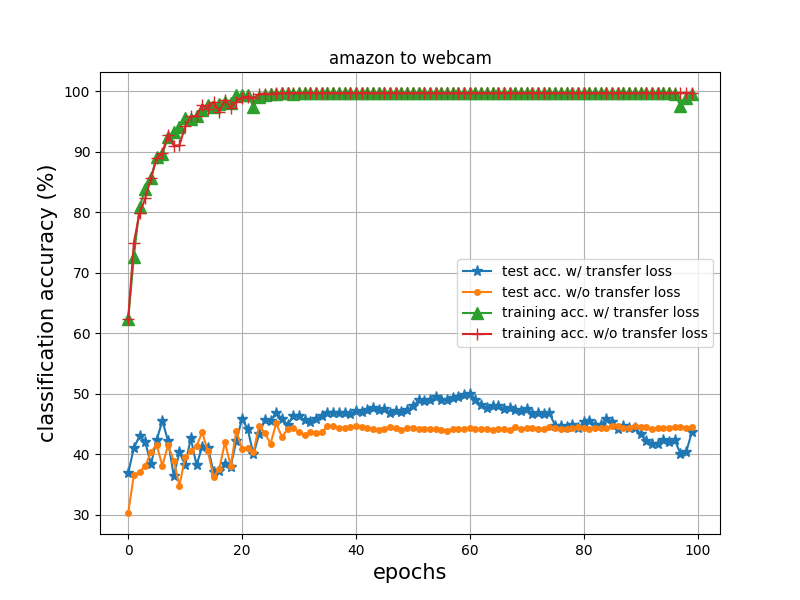}
\caption{CDAN}
\label{fig:cdan_acc}
\end{subfigure}%
\begin{subfigure}{.5\linewidth}
\centering
\includegraphics[scale=0.25]{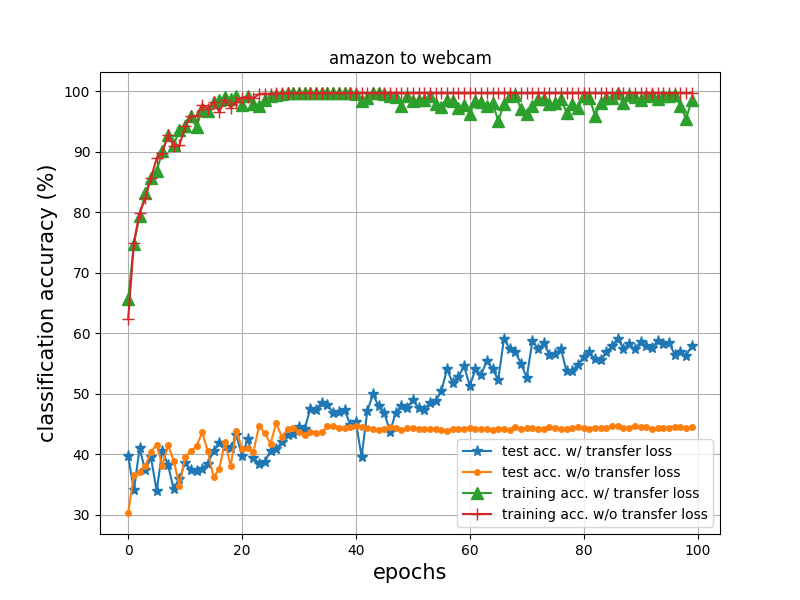}
\caption{CDAN+E}
\label{fig:cdan_e_acc}
\end{subfigure}\\[1ex]
\caption{Classification and transfer loss for each of the evaluated methods as a function of epochs.}
\label{fig:test}
\end{figure}

\begin{figure}[H]
\begin{subfigure}{.5\linewidth}
\centering
\includegraphics[scale=0.25]{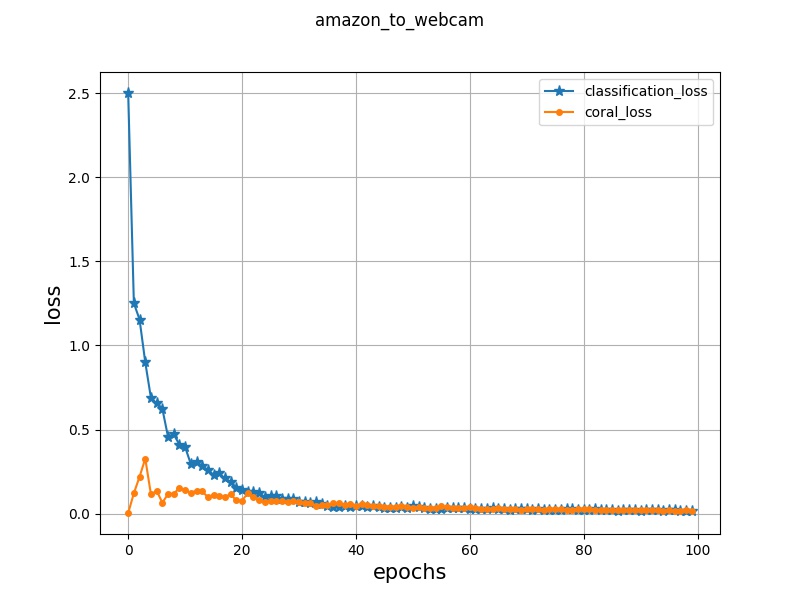}
\caption{DeepCORAL}
\label{fig:deepcoral_loss}
\end{subfigure}%
\begin{subfigure}{.5\linewidth}
\centering
\includegraphics[scale=0.25]{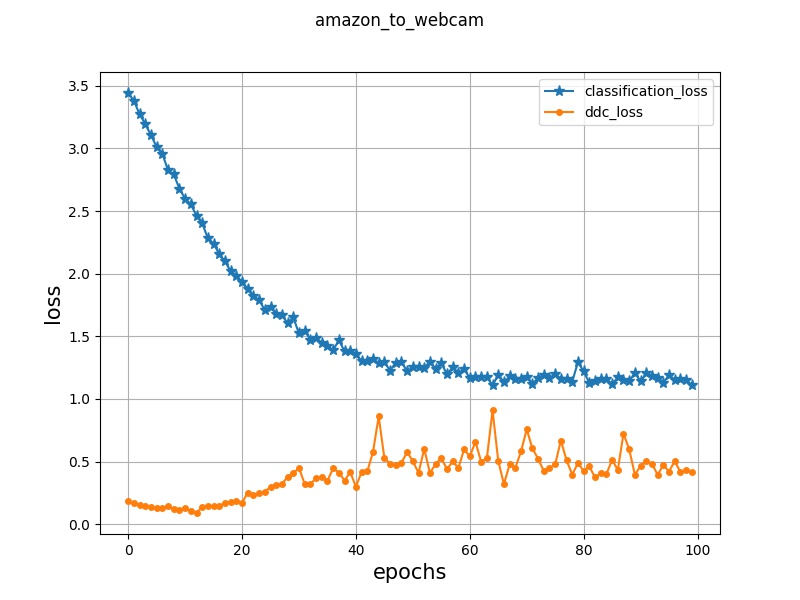}
\caption{DDC}
\label{fig:ddc_loss}
\end{subfigure}\\[1ex]
\begin{subfigure}{.5\linewidth}
\centering
\includegraphics[scale=0.25]{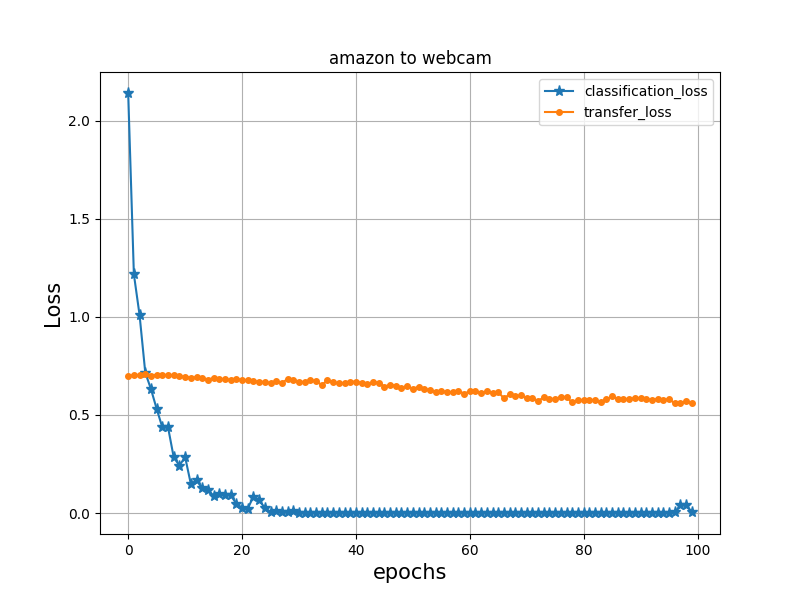}
\caption{CDAN}
\label{fig:cdan_loss}
\end{subfigure}%
\begin{subfigure}{.5\linewidth}
\centering
\includegraphics[scale=0.25]{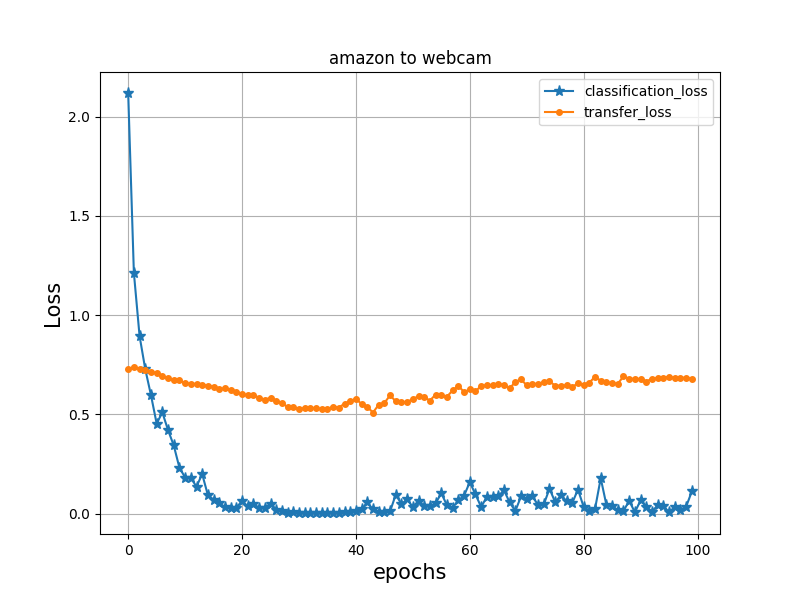}
\caption{CDAN+E}
\label{fig:cdan_e_loss}
\end{subfigure}\\[1ex]
\caption{Image recognition accuracies for the four evaluated methods. The domain shift in this case corresponds to Amazon to Webcam. Red and green lines correspond to the classification accuracy on the source test data. Blue and orange lines correspond to the target test data.}
\label{fig:test}
\end{figure}

As we obtained features f and predictions g, classification loss for source data can be calculated using $g_s$ and ground truth labels of source data. Transfer loss is calculated based on domain adaptation method selected. If CDAN is selected, joint variable h=(f,g) which have both source and target data features and predictions are used. Thus transfer loss is calculated using domain discriminator network whose input is h which is explained in above Sections on CDAN. If CDAN+E is selected, along with the features and predictions we calculated entropy from class predictions which is explained in above Section. CDAN+E \ref{sec:cdan+e}. This calculated entropy is used to perform weighted operations based on uncertainty of the predictions. Total loss to do back propagation is calculated using below equation.
\begin{equation}
    Total \ loss = classification \ loss + \lambda \ transfer \ loss
\end{equation}

where $\lambda$ is a hyperparameter which is described in the above Sections. During the experiment it is observed that transfer loss obtained is very small which resulted in very very small gradients of order $10^{-7}$. Which in turn did not update the weights of domain discriminator and thus the transfer loss is fluctuated with very small variance. In Table \ref{tab:accuracy_table} we can observe that CDAN+E only performed better in 2 cases out of 6. It is also observed that transfer loss is reducing till $20^{th}$epoch in most of the cases. In Fig. \ref{fig:cdan_acc} and \ref{fig:cdan_e_acc}, training and testing accuracy as well as  in Fig. \ref{fig:cdan_loss} and \ref{fig:cdan_e_loss}, transfer loss and classification loss of A$\to$W domain adaptation is shown respectively. In our opinion smaller batch size also played a key role in results of CDAN and CDAN+E methods. Smaller batch size is chosen considering the limitations of local machine. Although the results are not matched with original paper we observe that there is variation in accuracy and loss when transfer loss is included.

%%%%%%%%%%%%%%%%%%%%%%%%%%%%%%%%%%%%%%%%%%%%%%%%%%%%%%%
%\subsection*{Problems/Challenges encountered}
%- Understanding and implementing the end-to-end pipeline in Pytorch.
%
%- Data bottlenecks during training led us to use Google Colab or addressing a computer with higher memory capacity.
%
%- In Adversarial domain adaptation local machine is used and batch size is limited to 10.
%
%- In CDAN and CDAN+E, very small transfer loss is observed which resulted in very very small gradients of order $10^{-7}$. Which we believe resulted in vanishing gradient problem.
%
%- This resulted in poor results in both the methods and original implementation of authors of \cite{NIPS2018_7436} is also behaving the same way with same backbone and hyperparameters.
%%%%%%%%%%%%%%%%%%%%%%%%%%%%%%%%%%%%%%%%%%%%%%%%%%%%%%%

%%%%%%%%%%%%%%%%%%%%%%%%%%%%%%%%%%%%%%%%

\FloatBarrier
\begin{table*}[ht!]
\caption{Testing accuracy for different domain adaptation techniques across different datasets.}
\label{tab:accuracy_table}
\Large{
%\begin{tabular}{c|g|c|g|c|g|c}
\begin{tabularx}{\textwidth}{@{}lllllll@{}}
\toprule
              & A $\to$ W & A $\to$ D & W $\to$ A & W $\to$ D & D $\to$ A & D $\to$ W \\ \midrule
No Adaptation &     43.1$\pm$2.5      &   49.2$\pm$3.7        &    35.6$\pm$0.6       &       94.2$\pm$3.1    & 35.4$\pm$0.7  &   90.9$\pm$2.4
\\
DeepCORAL     &  \cellcolor{gray}  49.5$\pm$2.7       &   40.0$\pm$3.3        &       \cellcolor{gray} 38.3$\pm$0.4   &     74.4$\pm$4.3       &   \cellcolor{gray} {38.5$\pm$1.5}         &       \cellcolor{gray} 89.1$\pm$4.4     \\
DDC     &   41.7$\pm$9.1       &  ---         &     ---      &    ---       &       ---    &      ---          \\
CDAN          &    44.9$\pm$3.3       &   49.5$\pm$4.6        &   34.8$\pm$2.4        &  93.3$\pm$3.4         &   32.9$\pm$2.4        &     88.3$\pm$3.8      \\
CDAN+E        &    48.7$\pm$7.5       &     \cellcolor{gray} 53.7$\pm$4.7      &     35.3$\pm$2.7      &      \cellcolor{gray} 93.6$\pm$3.4   &    33.9$\pm$2.2       &    87.7$\pm$4.0       \\ \bottomrule
\end{tabularx}
}
\end{table*}
\FloatBarrier
%%%%%%%%%%%%%%%%%%%%%%%%%%%%%%%%%%%%%%%%

%%%%%%%%%%%%%%%%%%%%%%%%%%%%%%%%%%%%%%%
% - Add object recognition accuracies in table 
% - Add subplot with 4 graphs for accuracies (4 methods, same transfer domain)
% - Add subplot with 4 graphs for losses (4 methods, same transfer domain)
% \usepackage{booktabs}
%\newcolumntype{g}{>{\columncolor{Gray}}c}

%%%%%%%%%%%%%%%%%%%%%%%%%%%%%%%%%%%%%%%%
\section*{Conclusions}

Based on the methods we evaluated, we can say there are three possible scenarios for accuracy performance. Note that in the graphs presented here, the accuracy increases when we add a domain adaptation loss (e.g. Amazon $\,\to\,$ Webcam), however, we noticed that accuracy can decrease or stay the same for other domain shifts. Refer to the repository to see the other plots. Moreover, recognition accuracies don't go higher than 70\% in most experiments, this means there is a lot of room for improvement. For future work we can carry out trainings for 200 epochs (not 100) and understand better the behavior of our losses, for instance why in some cases losses oscillate too much. Finally, we can conclude that we gained hands-on experience building end-to-end pipelines using deep neural networks for domain adaptation tasks.

%%%%%%%%%%%%%%%%%%%%%%%%%%%%%%%%%%%%%%%%
\bibliographystyle{./bibliography/IEEEtran}
% \bibliography{./bibliography/IEEEabrv,./bibliography/IEEEexample}
% \bibliography{bibliography.bib}

\end{document}